\documentclass{article}

\usepackage{multirow}
\usepackage{arxiv}

\usepackage[utf8]{inputenc} % allow utf-8 input
\usepackage[T1]{fontenc}    % use 8-bit T1 fonts
\usepackage{hyperref}       % hyperlinks
\usepackage{url}            % simple URL typesetting
\usepackage{booktabs}       % professional-quality tables
\usepackage{amsfonts}       % blackboard math symbols
\usepackage{nicefrac}       % compact symbols for 1/2, etc.
\usepackage{microtype}      % microtypography
\usepackage{lipsum}		% Can be removed after putting your text content
\usepackage{graphicx}
\usepackage{natbib}
\usepackage{doi}

\date{}

\renewcommand{\headeright}{}
\renewcommand{\undertitle}{}

\begin{document}

\title{Attributing AUC-ROC to Analyze Binary Classifier Performance}

\author{
Arya Tafvizi \\
Google \\
aryat@google.com
\and
\textbf{Besim Avci} \\
Google \\
besim@google.com
\and
\textbf{Mukund Sundararajan}\\
Google \\
mukunds@google.com}

\maketitle

\begin{abstract}
Area Under the Receiver Operating Characteristic Curve (AUC-ROC) is a popular evaluation metric for binary classifiers. In this paper, we discuss techniques to segment the AUC-ROC along human-interpretable dimensions. AUC-ROC is not an additive/linear function over the data samples, therefore such segmenting the overall AUC-ROC is different from tabulating the AUC-ROC of data segments. To segment the overall AUC-ROC, we must first solve an \emph{attribution} problem to identify credit for individual examples. 

We observe that AUC-ROC, though non-linear over examples, is linear over \emph{pairs} of examples. This observation leads to a simple, efficient attribution technique for examples (example attributions), and for pairs of examples (pair attributions). We automatically slice these attributions using decision trees by making the tree predict the attributions; we use the notion of honest estimates along with a t-test to mitigate false discovery. 

Our experiments with the method show that an inferior model can outperform a superior model (trained to optimize a different training objective) on the inferior model's own training objective, a manifestation of Goodhart's Law. In contrast, AUC attributions enable a reasonable comparison. Example attributions can be used to slice this comparison. Pair attributions are used to categorize pairs of items---one positively labeled and one negatively---that the model has trouble separating. These categories identify the decision boundary of the classifier and the headroom to improve AUC.

% (b) Even when a complex model outdoes a simple one, the simple model can beat the complex one on some slices.  (c) Data-segments with good AUC-ROC nevertheless hurt the aggregate AUC-ROC. (d) Pair attributions can help us analyze a model's AUC-ROC headroom, and can also be used to study the model's decision boundary.
\end{abstract}




\section{Introduction}

\subsection{Why Segment?} 
Machine learning models are evaluated using a performance measure (e.g. loss or accuracy) on a held-out test set. This is a satisfactory evaluation under two conditions: (a) The model optimizes for an objective that measures the \emph{true} deployment objective (b) The test distribution is representative of the deployment scenario. 

In research/academic tasks (e.g. Imagenet~\cite{Imagenet} or Glue~\cite{glue}), these conditions are \emph{defacto} true; there is no notion of a deployment scenario that is separate from the training data distribution, and consequently is no deployment objective that differs from the training objective. 

However, in real world tasks, both conditions are always invalid to some degree. For instance, in a recommender system that recommends videos or music, while we may measure the quality of the recommender system based on whether the recommendation was accepted or not, or on time spent in engaging with the recommended content, these are only proxies for true user satisfaction. Likewise, the deployment distribution never matches the training distribution perfectly; there are always newer items that we don't know about, or seasonal variation in consumption patterns, or changes in taste, or strategic behavior from content creators in reaction to the recommendation system. 

To anticipate problems with deployment either due to non-representativeness or imperfect optimization objectives, it helps to segment the performance of the model. Indeed we are inspired by recent literature in this vein~\cite{Chung, Pastor2021}, which segment the evaluation data on linear metrics such as \emph{logloss} and \emph{false positive/negative rate} using decision trees and lattices. If a model's performance is variable across different segments, then it is less robust to distribution shifts and more sensitive to imperfections in the training objective. While it is hard to predict distribution shifts, it is easier to identify segments with variable performance, and have a decision-maker (e.g. a loan officer) judge whether segments with poor performance (say, student loans) are likely to increase in proportion. Moreover, identifying segments with outlying performance can help model builders make informed decisions on how/where to add more training data.

Furthermore, we often compare two models, one currently in use, and another a candidate to replace it. If one model is superior to another, it is worth checking if this superiority is robust to changes in the data distribution. Again, it makes sense to slice the difference in performance between the two models.

% \todobesim{More background on segmentation -- existing works, decision tree-based segmentation, what kind of metrics have been used for segmenting.}
% \todobesim{Practicality of this work.}
\subsection{Why AUC-ROC?}
\label{sec:why-auc}

In this paper, we focus on on a popular evaluation metric for binary classifiers called AUC-ROC (see Section~\ref{sec:prelim} for a definition). Let us justify this choice of metric.

Perhaps, the obvious choice of evaluation metric is the training objective itself. For classification tasks, for deep networks and logistic regression, this is typically cross entropy (CE) Loss (see Expression~\ref{eq:CE-Loss}. The overall CE Loss is a simple average of the CE Loss for each example, making this metric easy to decompose; for instance \cite{Chung} focuses on this metric. For random forests, it is an entropy based measure such as GINI Impurity (see Expression~\ref{eq:GINI-for-leaf}). Unfortunately, GINI Impurity does not apply to deep networks or logistic regression; so if you are comparing a random forest with a deep network, you cannot use GINI Impurity. You can use CE Loss, but as we discuss in Section~\ref{sec:misleading}, the results from such a comparison would be misleading because of a manifestation of Goodhart's Law; as we shall see, the logistic regression or deep learning would seemingly outperform random forests, but only because random forests don't target CE Loss. Furthermore, neither training objective is directly related to the model's classification performance, which is better captured by measures such as accuracy, precision and recall. 

Accuracy conflates type 1 (false positives) and type 2 errors (false negatives); when there is class imbalance, it is possible for a trivial classifier to outperform a non-trivial one. This is sometimes called the Accuracy Paradox. Using precision or recall requires us to commit to a threshold (for the score, above which the classifier returns `true'), i.e., a specific trade-off between the cost of false positives and false negatives. If a decision-maker has a specific trade-off in mind, these are good evaluation metrics; \cite{Pastor2021} focuses on these metrics. But often, it is hard to commit to a specific trade-off/threshold when we are evaluating a model.   

Then we are left with `global' metrics such as the F-score or AUC-ROC that are not linear over the examples. Neither metric can be easily segmented. We choose to work with the AUC-ROC because it is is closer to being linear (we make this precise in Section~\ref{sec:attribution-methodology}), and it does not require one to commit to a score threshold. A similar workflow for segmenting the F-score can be developed.

\subsection{Why Attribution?}

Because AUC-ROC is not linear, reporting the AUC-ROC for a list of segments is not the same as segmenting the aggregate AUC-ROC (see Example~\ref{tab:auc-example}); the aggregate AUC-ROC also includes contributions from the interaction between the segments. \emph{We seek to measure the contribution of a human-interpretable segment to the aggregate AUC-ROC, contributions that are on account of the quality of predictions and not just the number of examples in the segment. We seek to use this measure to automatically identify segments with very high or very low contributions to the aggregate AUC-ROC.} We will do this by first identifying the contribution of a single example to the aggregate AUC-ROC; this is the \emph{attribution} of the AUC-ROC to that example. We will then segment these attributions using standard tools such as decision trees.  

\subsection{Our Results}
\begin{itemize}
    \item In Section~\ref{sec:misleading}, we compare the performance of a random forest model and a logistic regression model on their training objectives. Unsurprisingly, each model handily beats (trounces) the other on its own training objective. We discuss that this is a manifestation of Goodhart's Law, i.e., `When a measure becomes a target, it ceases to be a good measure'. In contrast, comparing the aggregate AUC-ROC of the two models produces a more realistic picture that the two models perform similarly; AUC-ROC is almost never a training objective, i.e., never the `target' and therefore escapes the clutches of Goodhart's Law. This motivates us to segment AUC-ROC.
    \item In Section~\ref{sec:attribution-methodology}, we observe that though AUC-ROC is non-linear over examples, it is linear over pairs of examples. We use this observation to construct an efficient attribution technique for pairs of examples (\emph{pair attribution}) and for individual examples (\emph{example attribution}). In Section~\ref{sec:attribution-methodology}, we show that correctly normalized example attributions are nicely correlated with training objectives across examples, making them a good measure of model performance on an individual example.
    \item In Section~\ref{sec:segmentation} and Section~\ref{sec:analyzing-example-attributions}, we demonstrate the use of decision trees to segment AUC attributions. We use the notion of honest estimates with t-tests to mitigate false discovery. We show that segmenting by AUC attributions is as effective as segmenting by training objectives.   
    \item In Section~\ref{sec:comparing} we compare a simple forest and a regression model on AUC attributions; as discussed above, this is a safer comparison metric than using one of their objectives. We identify slices where the complex model truly outperforms the simple model.
    \item In Section~\ref{sec:headroom}, we segment pair attributions to study the headroom in AUC. We show that a model can have good AUC within a slice, and yet the slice can contribute to a low AUC via its interaction with other slices. 
    \item In Section~\ref{sec:decision-boundary}, Figure~\ref{fig:cross-auc-lending} we segment pair attributions, demonstrating how we can identify pairs of segments of examples, one consisting of positive examples and the other of negative examples, that the model finds hard to separate. 
\end{itemize}

\section{Related Work}
\label{sec:related}

\textbf{AUC-ROC:}
There is a literature on analyzing (aggregate) ROC graphs (see for instance~\cite{Fawcett}); they study properties like convexity of the ROC curve, isometry lines for accuracy, class skew sensitivity of ROC, and optimizing AUC-ROC. In contrast, we discuss how to identify the AUC-ROC contribution of an example, or a data segment. 

There is also work (e.g.~\cite{pmlr-v54-eban17a, yuan2021robust}) on optimizing AUC-ROC. To do this, the authors of \cite{pmlr-v54-eban17a} produce additively-separable, differentiable upper and lower bounds on true positives, true negatives, false positives and false negatives. \cite{yuan2021robust} proposes a new margin-based min-max surrogate loss function for the AUC score. In contrast, we use a simple attribution technique to distribute AUC-ROC credit to positive-negative example pairs in the evaluation dataset for a trained model.

\textbf{Segmentation:}
There is work on segmenting the performance of classifiers (e.g.~\cite{Chung, Pastor2021}).
\cite{Chung} describes techniques to slice data in order to identify subsets of examples where the model performs poorly. The major distinction is that they slice loss, an additive metric, whereas we slice AUC-ROC, a global, non-additive evaluation metric; indeed their segmentation technique based on a lattice search, could also be applied to our AUC attributions. A minor distinction is that in addition to a t-test (also employed by~\cite{Chung}), we use honest estimation to mitigate false discovery (see Section~\ref{sec:segmentation}).

Pastor et al. \cite{Pastor2021} aim to estimate the divergence on classification behavior in data subgroups w.r.t. overall behavior. The technique proposed in this paper works by finding frequent item sets (feature combinations) and then identifying the ones whose metric values are significantly different from the the whole dataset.
%However, this approach would suffer from Multiple Comparisons Problem \cite{benjamini1995} because of performing exponentially many statistical tests on all candidate slices.
This paper's main focus is on false positive (FPR) and false negative rates (FNR), commonly-used fairness metrics; In contrast we focus on AUC-ROC (justified in Section~\ref{sec:why-auc}).

% their technique supports other additive classification metrics such as accuracy, true positive rate, positive predictive value. Differently from \cite{Chung}, they also calculate Shapley-based contribution of each attribute to slice divergence. 
% Explain why slicing first and comparing the metrics later is not ideal
% correlating features will be a problem.
% The number of slices to explore should grow exponentially with feature cardinality, but the experiments show it doesn't.
% they could face false discovery.

\textbf{Other Evaluation Techniques:}
There is other work on addressing deficiencies in the held-out test metric. For instance~\cite{ribeiro-etal-2020-beyond}, discusses the use of a test framework to validate linguistic capabilities in NLP models. There is also progress in making classifiers aware of out-of-distribution examples (e.g.~\cite{meinke2021provably, BitterwolfM020}). In contrast, we are interested in identifying variable performance on in-distribution examples. 

\textbf{Attribution:}
There is literature on attributing prediction scores to model features (e.g.~\cite{STY17,Lundberg2017AUA} or to training examples~\cite{tracin}. These attribution methods contend with non-linearities in machine learning models. In contrast, we contend with non-linearities in evaluation metrics (AUC-ROC).

% \todobesim{Discuss a qualitative comparison to the Shapley value. (a) undefined for sets with only one polarity. Perhaps if you have one side of the bipartite graph present and the other appear online?}

% \todobesim{Discuss Fred's point about slicing accuracy by using various thresholds. AUC then defines a specific speed for the threshold change.}

% \todobesim{cite: Robust Deep AUC Maximization: A New Surrogate Loss and Empirical Studies on Medical Image Classification. Their technique in section 3.3 is similar to how we think about AUc.}

\section{Preliminaries}
\label{sec:prelim}
We study machine learning (ML) evaluation. It is common practice in ML to evaluate the model using an evaluation metric on a random sample of data that is \emph{held out} of the training process; we will call this set the \emph{test} set. All of our empirical results are on randomly sampled test sets.

We discuss standard classifier evaluation metrics. The first two metrics, cross entropy loss (CE Loss) and GINI impurity (GINI) are standard training objectives, CE Loss for logistic regression and deep learning, and GINI Impurity for random forest classifiers.

\subsection{Cross-Entropy Loss}
CE is used to train deep learning models and logistic regression models. CE Loss takes as inputs prediction and labels for all examples in the test set. CE Loss for the test set is the sum of CE Loss for individual examples. For binary classification, CE Loss for an individual example with a prediction probability $\hat{y}$ for the `true' class and a label $y \in \{0,1\}$  is: 

\begin{equation}
\label{eq:CE-Loss}
    - y \log(\hat{y}) + (1-y) \log(1- \hat{y}) 
\end{equation}

The Cross Entropy Loss for a dataset is the average of the losses for the examples in the data set. It is also the metric used for segmenting in ~\cite{Chung}.

% For Multiclass classification, this generalizes to:

% \begin{equation}
%     \sum_{c} - y_c \log(\hat{y_c}) 
% \end{equation}

\subsection{Gini Impurity}
While GINI is not a common evaluation metric, we present it here for symmetry and to support Section \ref{sec:misleading}.
GINI is used to train classification trees and random forests. Forests are a collection of trees, where the prediction of the forests is the average of the predictions from the trees. 
To classify an example, a tree routes the example to a leaf node. GINI impurity is a measure of entropy of the set of examples that are routed to the same leaf. For a single leaf, suppose $p_c$ is the fraction of examples from class $c$ at that leaf. Then, GINI impurity is defined as: 

\begin{equation}
\label{eq:GINI-for-leaf}
    \sum_{c} p_c (1 - p_c) 
\end{equation}

One way to interpret this definition is that the impurity of an example with class $c$ is $1-p_c$, and the impurity of the leaf is the expected impurity from sampling an example from a leaf. The GINI impurity for a tree is the weighted sum of leaf impurities, where a leaf's weight is the fraction of examples that route to the leaf. \footnote{With a bagging approach, this is the fraction of examples from the random sample used to construct the tree.} The impurity of a forest is a simple average of the trees in the forest. \footnote{There are a couple of ways to compute the impurity of the test set. One approach is to use the structure of the trees to identify the leaf sets, but to use the counts from test set to define the associated probabilities; this would ignore the scores we get from training. The other approach is to define the impurity of a test example as $1-p$, where $p$ is the probability score for the ground-truth label. The impurity of the test set is then the average of the examples in the test set. We take the second approach to remain faithful to the scores returned by the classifier. This approach also allows us to measure the impurity for logistic regression, though it does not have an underlying notion of sets or entropy. While we sometimes report GINI Impurity for Logistic Regression, we do not defend this.} 

\subsection{AUC-ROC}
The other evaluation metric that we discuss is AUC-ROC (henceforth shortened to AUC), which is not used for training, but only in evaluation. The AUC is defined as a function of the True Positive Rate (TPR), i.e., the fraction of true labels that are correctly classified as true and the False Positive Rate (FPR), i.e., the probability that a classifier is incorrect when it labels an example as true. When a classifier returns a probability score for the true label, just as random forests, logistic regression and deep learning models all do, it is possible to vary the threshold above which we classify an example as true. High thresholds have a low TPR (bad) and low FPR (good), and low thresholds tend to have high TPR (good) and high FPR (bad). This sets up a trade-off. If we plot TPR as a function of FPR, we get a trade-off curve. AUC is the area under this curve, it is a measure of the model's goodness agnostic of a specific threshold. It ranges from $0$ to $1$. Better classifiers have higher scores.

\subsection{Data sets and Tasks For Experiments}
\label{sec:datasets}

For the experiments in this paper, we train a logistic regression and a random forest model on two datasets. The first dataset from the UCI Machine Learning Repository~\cite{UCI}; the data is drawn from the US census, and the task is to predict whether an adult's income exceeds 50K based on features of the individual and their household. The positive class corresponds to adults with incomes that exceed 50K. This dataset contains around 33 thousand examples and 13 features.
The second dataset is a dataset~\cite{lending} from the Lending Club~\cite{lending2}, a platform for peer to peer lending. The task is to predict if a customer will default on their loan. The positive class corresponds to customers who default (despite the fact that this is a `negative' event). This dataset contains around 1.4 million examples and 23 features.

\section{Misleading Results from Comparing Models on Training Objectives}
\label{sec:misleading}

We compare the performance of a logistic regression model and a random forest model, both trained and evaluated on the same split of the Census dataset in Table~\ref{tab:loss-example-census}. Each model significantly outperforms the other model when evaluated on the metric it was trained to optimize. The logistic regression has $~25\%$ lower CE loss (the objective of logistic regression) than the random forest. The random forest has $~20\%$ lower impurity (the objective of random forests) than the logistic regression model. This phenomena is not explained by over-fitting; all the evaluation is done on held out test sets. Which comparison should we trust?

 The answer is neither. This is Strathern's (\cite{strathern1997improving}) version of Goodhart's Law~(\cite{Goodhart1984}) in action. The statement of the law is: `When a measure becomes a target, it ceases to be a good measure'. 
 
 Here, the problem surfaces because each `measure' is a `target' for one model, but not the other. CE Loss is a target for the logistic regression model but not the random forest, and GINI Impurity is a target for the random forest model, but not the linear regression.
 
 The table shows that phenomenon mostly persists across age segments; slicing does not clarify which model is superior. 

In contrast, AUC comparison shows that both models have similar performance (the random forest is marginally better).

\begin{table*}[]
\centering
\begin{tabular}{|l|l|l|l|l|l|l|}
\hline
\textbf{Slice}     & \textbf{CE Loss RF} & \textbf{CE Loss LR} & \textbf{GINI RF}      & \textbf{GINI LR}  & \textbf{AUC RF} & \textbf{AUC LR}      \\ \hline
All data  & 0.48       &   0.36     & 0.19         & 0.24     &   0.89  & 0.88             \\ \hline
Age<=25   & 0.07       &   0.09     & 0.03         & 0.07     &   0.97  & 0.97           \\ \hline
25<Age    & 0.72       &   0.42     & 0.23         & 0.28     &   0.87  & 0.85            \\ \hline
25<Age<34 & 0.62       &   0.34     & 0.18         & 0.23     &   0.86  & 0.85        \\ \hline
\end{tabular}
\caption{A logistic regression model (LR) and a random forest model (RF) trained on the same data and features are evaluated on different Age segments. Each model outperforms the other when evaluated on the metric it was trained to optimize. AUC comparison shows that both models have similar performance, with the random forest model performing marginally better.\label{tab:loss-example-census}}
\end{table*}

\section{Attribution to Examples}
\label{sec:example-attributions}
The phenomenon in Section~\ref{sec:misleading} motivates us to compare and slice model performance on a metric that is not used as a training objective; this is true of AUC. In Section~\ref{sec:auc-is-not-linear}, we show that AUC is not linear motivating the attribution approach in Section~\ref{sec:attribution-methodology}. We combine this with the segmentation methodology in Section~\ref{sec:segmentation} to segment the performance of a single model in Section~\ref{sec:analyzing-example-attributions}, and to compare two models in Section~\ref{sec:comparing}.

\subsection{AUC is not Linear}
\label{sec:auc-is-not-linear}

The example in Table~\ref{tab:auc-example} shows us that unlike GINI Impurity and CE Loss \footnote{With CE Loss each example contributes the quantity in Expression~\ref{eq:CE-Loss} divided by $m$, the number of training examples. For GINI Impurity, each example contributes $\frac{1-p}{m}$, where $p$ is the probability score for the ground truth label. }, AUC is a non-linear metric.

To make sense of the example, let us first discuss a well-known redefinition of AUC: AUC can be defined as the fraction of pairs of examples, one with a positive label, and the other with a negative label, that are \emph{ordered correctly}; a pair is ordered correctly if the score of the example with the positive label exceeds the score of the example with the negative label. (Ties contribute $1/2$.) Using this, we can compute the AUC of the dataset in Table~\ref{tab:auc-example} to be $8/9$ because only one of the $9$ pairs of examples (pairs that have different labels) is incorrectly ordered. 

 Now suppose that the dataset can be grouped into three slices $A,B,C$ with two elements each as shown in the table; the AUCs for the slices are $1,0,1$ respectively. The average of the three AUCs is $2/3 \neq 8/9$. The reason is that the overall AUC computation involves pairs of examples \emph{across} the slices; averaging over slice AUC ignores this. 
 
 We can also show that improving the AUC for a slice, can hurt overall AUC. For instance if the third example has a score of $0.6$ and the fourth example has a score of $0.7$, then the AUC of slice B is $1$, but the overall AUC drops to $7/9$.
 
 This discussion demonstrates why displaying the AUC of some segments is not the same as segmenting the aggregate AUC; for the latter, we have to solve an \emph{attribution} problem, i.e., identify the contribution of an individual example to the aggregate AUC.  

\begin{table}[]
\centering
\begin{tabular}{|l|l|l|l|l|}
\hline
 & \textbf{Label} & \textbf{Pred} & \textbf{Slice}              & \textbf{AUC} \\ \hline
1       & 0     & 0.1          & \multirow{2}{*}{\textbf{A}} & \multirow{2}{*}{1} \\ \cline{1-3}
2       & 1     & 0.5          &                             &                    \\ \hline
3       & 0     & 0.3          & \multirow{2}{*}{\textbf{B}} & \multirow{2}{*}{0} \\ \cline{1-3}
4       & 1     & 0.2          &                             &                    \\ \hline
5       & 0     & 0.1          & \multirow{2}{*}{\textbf{C}} & \multirow{2}{*}{1} \\ \cline{1-3}
6       & 1     & 0.5          &                             &                    \\ \hline
\end{tabular}
\caption{\label{tab:auc-example} A data set with six examples and three slices that shows that AUC is not linear. Overall AUC is $8/9$, which is not the average of slice AUCs ($2/3$).}
\end{table}

\subsection{Attribution Methodology}
\label{sec:attribution-methodology}

While AUC is non-linear, it is only a second-degree set function in the sense that it depends only on subsets of size $2$, i.e., pairs of elements. We will take two approaches to analyzing AUC in this paper. 

\subsubsection{Pair Attribution}
\label{sec:pair-attribution-methodology}
In the first approach which we call  \emph{pair-attribution}, we will attribute AUC to pairs, i.e., each correctly ordered pair gets a credit of $1$ and each incorrectly ordered pair gets a credit of zero. In this sense we are redistributing $p*n*AUC$, a quantity also known as the U statistic, where $p/n$ are the number of positive/negative examples in the set. We will apply this methodology to study the headroom in AUC and the decision boundary of a classifier in Section~\ref{sec:pair-attributions}.

\subsubsection{Example Attribution}
\label{sec:example-attribution-methodology}
In the second approach we call \emph{example attribution}, for each correctly ordered pair, we assign a credit of $1/2$ to each of the examples in the pair. For each incorrectly ordered pair, both examples get a credit of zero. The \emph{total} attribution for an example sums over all the pairs that it participates in. If we divide the total attribution by $p*n$, and sum the this over all examples, then we will get the aggregate AUC. 

We will also compute \emph{normalized} example attributions, by \emph{averaging} the credit over all the pairs an example participates in, instead of summing. This corresponds to dividing the total attribution for a positive example by $n$ and the total attribution for a negative example by $p$. 

Notice that example and pair attributions can both be computed in time quadratic ($n*p$) in the dataset size. (This is cheaper and simpler than attribution techniques such as Shapley values that are based on permutations of the examples.) For large datasets, it is possible to sample the examples, compute the attributions over samples, and then segment the sample.
 
\begin{table*}[]
\centering
\begin{tabular}{|l|l|l|l|l|}
\hline
\textbf{Correlation}               & \textbf{CE Loss RF} & \textbf{CE Loss LR} & \textbf{GINI Impurity RF} & \textbf{GINI Impurity LR} \\ \hline
AUC Attribution           &   -0.06   &    +0.22    &       +0.14      &     +0.28      \\ \hline
Normalized AUC Attribution &   -0.53   &    -0.72    &       -0.84      &     -0.76   \\ \hline   
\end{tabular}
\begin{tabular}{|l|l|l|l|l|}
\hline
\textbf{Correlation}               & \textbf{CE Loss RF} & \textbf{CE Loss LR} & \textbf{GINI Impurity RF} & \textbf{GINI Impurity LR} \\ \hline
AUC Attribution           &   +0.19   &   +0.53    &     +0.44      &   +0.59               \\ \hline
Normalized AUC Attribution &    -0.32  &  -0.29     &     -0.54      &       -0.32           \\ \hline   
\end{tabular}
\caption{Normalized AUC attributions have strong correlation (in the right direction) with both CE Loss and GINI Impurity. Unnormalized AUC attributions either have weak correlation, or correlation in the wrong direction. (top: Census data, bottom: Lending data)\label{tab:loss-correlation}}
\end{table*}

We now show that normalized example attributions can be used as a measure of the performance of the model on an individual example. To do this, we study the Pearson correlation of the AUC attributions versus the model's own training objective, which we know to be a satisfactory measure of model performance on an individual example (unless it is used to perform apples-to-oranges comparisons as discussed in Section~\ref{sec:misleading}). Recall that polarity of AUC is the opposite to that of GINI and CE Loss; \emph{high} AUC (and hence high AUC Attribution) is good, whereas low GINI impurity and low CE loss are good. Therefore, if AUC is a good measure, we would expect a high \emph{negative} correlation.       

In Table~\ref{tab:loss-correlation} we show results for AUC attributions with and without normalization (see Section~\ref{sec:attribution-methodology})). We find that the (unnormalized) AUC attributions have poor correlation, or correlation in the wrong direction; this is possibly because these quantities are proportional to the number of pairs an example participates in. In contrast, we find that the \emph{normalized} AUC attributions have a strong correlation in the expected (negative) direction. Hereafter, we use the normalized AUC attributions for segmentation. 

\subsection{Segmentation}
\label{sec:segmentation}

We would like to automatically identify interpretable slices which have extreme contributions to the overall AUC, either very high or very low. To do, this we will segment the AUC along a list of dimensions specified by the analyst. To automate the segmentation process, we will reduce the segmentation problem to a prediction problem, i.e., we will use a decision tree with the specified dimensions as the features to \emph{predict} the normalized AUC attribution of the examples. The leaves of this tree consist of data segments with consistent contributions within the slice, and those slices with either very high or very low contributions are of particular interest. We report the normalized AUC attribution averaged over the examples in each leaf.

If the data at a leaf node is sparse, it could result in a noisy estimate of the average attribution of a segment. In a decision tree, this noise influences leaf estimates, but it also affects the splits, and the splits determine the reported segments, resulting in false discovery. To mitigate this, we use the notion of honest estimates from~\cite{Athey7353}. The idea is to keep part of the test aside and train the tree on the rest. Finally, we apply a t-test to check if the examples in the leaf segment from the two test samples are drawn from the same distribution (this is a good segment) or not (this is a false discovery); we constrain leaves to have a size of at least a $100$ to make the estimate of the mean normally distributed for a valid t-test. In tables we mark segments that constitute a false discovery with an asterisk; e.g. see the second-last row of Table~\ref{table:auc-segment-random-forests}. In our segmentation trees, we color these segments a darker shade; e.g. see the fifth leaf from the left in Figure~\ref{fig:pair-segment-lending}.

\subsection{Analyzing Example Attributions}
\label{sec:analyzing-example-attributions}

The normalized AUC attributions can reliably be used to identify segments of data where a model performs better or worse than overall. We show that segmenting on the mean AUC Attribution produces similar results from segmenting on the model's training objective. 

On the Census dataset, mean AUC attribution on predictions from the logistic regression model is ~10\% lower in the \emph{Married-civ-Spouse} segment compared to all data, and is ~10\% higher on the complement of that segment. Similarly, Mean AUC Attribution on predictions from the random forest model diverges ~7\% from all data on that segment and its complement.

Mean CE Loss on predictions from the logistic regression model, and mean GINI Impurity from on predictions from the random forest model are both ~50\% higher on the \emph{Married-civ-Spouse} segment and ~50\% lower on its complement. 

Recall from Section~\ref{sec:misleading}, that we should expect larger swings for the training-objective compared to the Mean AUC attributions because the training objective is the `target'. 

We observe the same pattern on the Lending dataset, where mean AUC attributions diverge from all data by ~10\% on predictions from the logistic regression, and by ~20\% on predictions from the random forest. Mean CE Loss diverges ~20\% on the same segments with predictions from the logistic regression, and mean GINI Impurity ~40\% with predictions from the random forest.

These patterns together confirm that the correlations between AUC attributions, CE Loss and GINI Impurity on data points, discussed in Section~\ref{sec:attribution-methodology}, propagate through to large segments of data. Insights from studying metric values on segments are similar for these metrics.

Table~\ref{table:auc-segment-random-forests} and Table~\ref{tab:auc-segment-lending} report AUC attributions on segments with different model performance, as discovered by a decision tree. There is a variation in performance of about $10\%$ for both tasks across slices. If the deployment scenario happens to contain a higher proportion of examples from slices with bad AUC attribution, the test AUC will be an overestimate of the real performance.

\subsection{Comparing Attributions across Models}
\label{sec:comparing}

\begin{figure*}[h]
\caption{A decision tree identifies segments in the Census dataset where the logistic regression and the simple random forest receive different AUC attributions.}
\centering
\includegraphics[width=\textwidth]{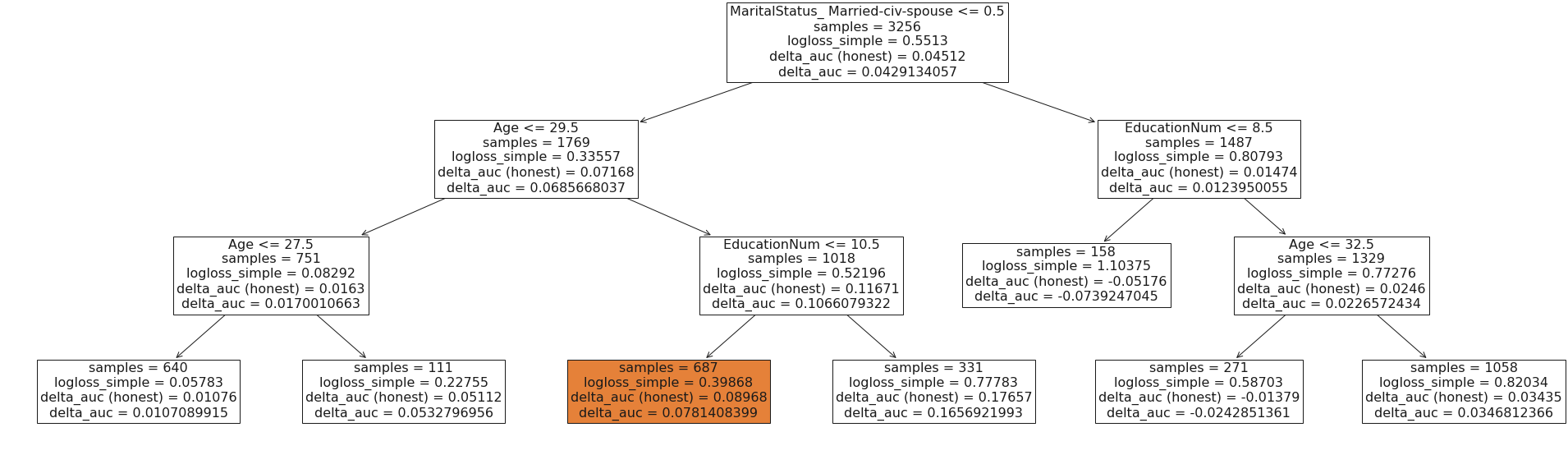}
\label{fig:delta-tree}
\end{figure*}

Section~\ref{sec:misleading} highlighted the risks of evaluating a model on its  training objective. Now we use AUC to compare a logistic regression model (trained on 13 features, some  numerical, some categorical) and a very simple random forest with access to only \textit{Age} and \textit{EducationNum} features for the Adult Income task. This comparison discovers areas where the added complexity due to extra feature interactions may be warranted. Recall that the two models are trained on different training objectives, CE Loss for the logistic regression, and GINI Impurity for the random forest. Therefore, as discussed in Section~\ref{sec:misleading}, we should not compare the two models on either training objective, justifying the focus on AUC-ROC.

In this analysis, we modify the segmentation approach from Section~\ref{sec:segmentation}, to predict the \emph{difference} in the mean AUC attributions of the two models. The aggregate AUC for the simple (random forest) model is $0.79$, and for the larger logistic regression is $0.88$, so there is a gain of $9\%$. Let us analyze where this gain is coming from. 

As seen in Figure~\ref{fig:delta-tree}, the simple model performs nearly as well, or better than the larger model on some of the leaves, but performs much worse on the fourth leaf, which is a slice of unmarried individuals older than 29.5 years and with more than 10.5 years of education. In the fifth leaf, the segment of married individuals younger than 32.5 years with more than 8.5 years of education, the simple model outperforms the larger model! 

Table~\ref{tab:auc-delta-segment-lending} repeats this analysis on the Lending dataset.

Imagine that the simple model is the current model in production, and the complex model is a candidate to replace it, then we can make a more informed launch decision based on where the wins and losses are coming from.

% nearly as well as the logistic regression model on the \textit{Married-civ-Spouse} slice, demonstrating that using a larger model does not significantly improve the performance on this slice.  
% However, the larger model performs better on the complement slice and on sub-slices defined by interactions with other features. The obvious explanation for the improvement is that the larger model includes \textit{Marital Status}, and therefore can condition on Marital Status.

\begin{table}[]
\centering
\begin{tabular}{|l|l|l|l|}
\hline
\textbf{Slice}                           & \textbf{AUC Attribution}   \\ \hline
All data                        &       0.45                       \\ \hline
not Married-civ-Spouse          &       0.47             \\ \hline   
Married-civ-Spouse              &       0.41             \\ \hline  
not Married-civ-Spouse, EducationNum<=10.5  &       0.48                             \\ \hline
not Married-civ-Spouse, EducationNum>10.5  &       0.45                             \\ \hline
Married-civ-Spouse, HoursPerWeek<=41.5  &       0.41*                             \\ \hline
Married-civ-Spouse, HoursPerWeek>41.5  &       0.42                             \\ \hline
\end{tabular}
\caption{A decision tree is used to identify segments in the Census dataset with large divergence of AUC contributions for the random forest model. Honest estimates are reported. * marks estimates which are noisy.} \label{table:auc-segment-random-forests}
\end{table}

\begin{table}[]
\centering
\begin{tabular}{|l|l|l|l|}
\hline
\textbf{Slice}                           & \textbf{AUC Attribution}   \\ \hline
All data                        &       0.35           \\ \hline
int\_rate<=12.5                  &       0.41             \\ \hline   
int\_rate>12.5                   &       0.31             \\ \hline  
int\_rate<=12.5, not grade\_A    &       0.38             \\ \hline   
int\_rate<=12.5, grade\_A       &       0.45             \\ \hline   
15.4>int\_rate>12.5           &       0.33             \\ \hline  
int\_rate>15.4                  &       0.29             \\ \hline  

\end{tabular}
\caption{A decision tree is used to identify segments in the Lending dataset with large divergence of AUC attributions to predictions from a random forest model. Honest estimates are reported. \label{tab:auc-segment-lending}}
\end{table}

% \begin{table}[]
% \begin{tabular}{|l|l|l|l|}
% \hline
% \textbf{Slice}                           & \textbf{Delta AUC Attribution}   \\ \hline
% All data                        &       0.04                       \\ \hline
% not Married-civ-Spouse          &       0.07             \\ \hline   
% Married-civ-Spouse              &       0.01             \\ \hline  
% not Married-civ-Spouse, Age>29.5  &       0.11                             \\ \hline
% not Married-civ-Spouse, Age<=29.5  &       0.02                             \\ \hline
% Married-civ-Spouse, Age<=37.5  &       0.01                             \\ \hline
% Married-civ-Spouse, Age>37.5  &       0.03*                             \\ \hline
% \end{tabular}
% \caption{Segments with relatively large difference in normalized AUC attribution between two models - a logistic regression model trained on all features, and a simple random forest model trained on only \textit{Age}, and \textit{EducationNum} features - are identified.\label{tab:loss-delta}}
% \end{table}

\begin{table}[]
\centering
\begin{tabular}{|l|l|l|l|}
\hline
\textbf{Slice}                           & \textbf{Delta AUC Attribution}   \\ \hline
All data                        &       0.06                       \\ \hline
int\_rate<=21.3             &       0.06             \\ \hline   
int\_rate>21.3              &       0.12             \\ \hline  
int\_rate<=7.65             &       0.04             \\ \hline   
int\_rate<=21.3, interest rate>7.65&0.06             \\ \hline   
int\_rate>21.3, fico\_range\_high<=701.5 &       0.09             \\ \hline  
int\_rate>21.3, fico\_range\_high<=701.5 &       0.25             \\ \hline  

\end{tabular}
\caption{Segments with relatively large difference in normalized AUC attribution between two models - a logistic regression model trained on all features, and a simple random forest model trained on only \textit{int\_rate}, \textit{loan\_amnt}, and \textit{purpose} features - are identified.}
\label{tab:auc-delta-segment-lending}
\end{table}

\section{Attribution to Pairs of Examples}
\label{sec:pair-attributions}
In this section we apply the pair attribution methodology (see Section~\ref{sec:pair-attribution-methodology}) to analyze the headroom in AUC (Section~\ref{sec:headroom}) and the decision boundary of a logistic regression classifier (Section~\ref{sec:decision-boundary}).

\begin{table*}[]
\centering
    \begin{tabular}{|l|l|c|}
        \hline
        \textbf{Negative Slice (Income <= 50K)}                     & \textbf{Positive Slice (Income > 50K)}                                & \textbf{AUC Attribution} \\ \hline
        MaritalStatus = Married                    & EducationNum $<= 10.5$ and CapitalGain $<= 5056$ & 0.430            \\ \hline
        MaritalStatus = Married                    & EducationNum $> 10.5$ and MaritalStatus != Married & 0.470            \\ \hline
        MaritalStatus != Married and CapitalGain $<= 4533$ & MaritalStatus != Married and EducationNum $<= 12.5$   & 0.918           \\ \hline
        MaritalStatus != Married and CapitalGain $<= 4533$ & MaritalStatus = Married               & 0.992           \\ \hline
    \end{tabular}
\caption{Selected segments that have substantially high or low AUC attributions compared to other segments from the Census model. Note that `Married-civ-spouse' is shortened to `Married.'}
\label{table:pair_slices}
\end{table*}

\subsection{Grouping Pairs}
In Sections~\ref{sec:normalized-aggregation} and~\ref{sec:headroom}, we will group pair attributions over a partition of the data $\{A_i\}$, by grouping pairs over the the Cartesian product of the sets $A_i$. Any pair of examples $p,q$ belong to a cross $A_i \times A_j$ if $p$ is positively labeled, and $q$ is negatively labeled and $p$ belongs to segment $A_i$ and $q$ belongs to segment $A_j$.

\subsection{Segmenting AUC along Specified Dimensions}
\label{sec:normalized-aggregation}

In this section, the score for a cross $A_i \times A_j$ is the \emph{fraction} of pairs belonging to the cross that are correctly ordered, i.e., the score for the positively labeled example exceeds the score for the negatively labeled example in the pair. This is the \emph{mean} of the pair attributions for pairs of examples belonging to the cross; we call this the \emph{mean pair attribution} for a cross. 

\begin{figure}[h]
\caption{AUC computed for the logistic regression model on \textit{Loan Grade} slices in the Lending dataset. Diagonal entries correspond to the fraction of correctly ordered pairs where both items belong to the same category. The off-diagonal entries similarly correspond to pairs where the items belong to different categories.}
\label{fig:cross-auc-lending}
\centering
\includegraphics[width=0.40\textwidth]{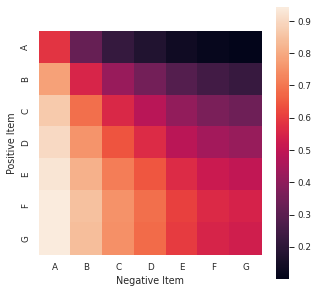}
\end{figure}

Consequently the mean pair attribution for the cross $A_i \times A_i$ is just the AUC of the slice. In Figure~\ref{fig:cross-auc-lending}, we will plot the score for a cross $A_i \times A_j$ in the cell corresponding to row $i$ and column $j$. 

Figure~\ref{fig:cross-auc-lending} describes the heterogeneity in the logistic regression model's AUC performance across partitions of the data along the \emph{Loan Grade} dimension. In the Lending Club dataset (see Section~\ref{sec:datasets}), Loan Grade is a hand-tuned formula that combines credit scores with other indicators of credit risk from the credit report and the loan application. This explains the smooth shading that we see in the figure; The model finds it easy to separate positive items (defaults) from grade G and negative items (no default) from Grade A; we expect loans from grade G to default at a high rate, and loans from grade A not to default. In contrast, the model finds it hard to separate negative items from G and positive items from A; this is a counter-intuitive phenomenon, and unsurprisingly, it is hard to model.   

% That is, the model is able to separate positive and negative items to a different degree depending on the category to which each item in the pair belongs. For instance $Wife=Positive x Own Child=Negative$ has a comparatively high AUC. In contrast, $Own Child=Positive X Wife=Negative$ has comparatively low AUC, and $Wife=Positive x Wife=Negative$ and $Own Child=Positive X Own Child=Negative$ have middling AUCs.

\begin{figure*}[]
\caption{Tree-based segmentation of downsampled (to 10K) pairwise Lending Club data. Features of negative examples are suffixed by `\_neg' and positive examples by `\_pos'. Minimum leaf size was set to 1000.}
\label{fig:pair-segment-lending}
\centering
\includegraphics[width=\textwidth]{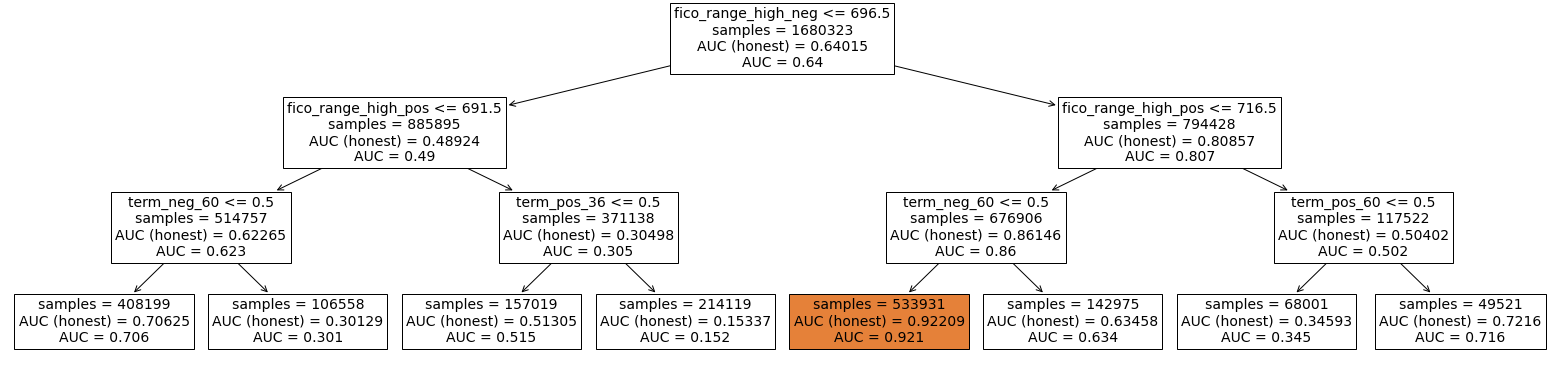}
\end{figure*}

This analysis can be seen as a study of the \emph{decision boundary} of a classifier, i.e., what type of pairs of examples, one positive and one negative, can it separate easily, and what type of pairs does it find hard to separate.

\subsection{Investigating Headroom in Aggregate AUC}
\label{sec:headroom}

Recall from the discussion in Section~\ref{sec:auc-is-not-linear} that a model has a perfect AUC of $1$ if all pairs of examples, consisting of one positively labeled example and one negatively labeled example, are `ordered correctly', i.e., the score for the positive exceeds the score for the negative. In this section, we categorize the \emph{headroom} in improving AUC, i.e., we categorize incorrectly ordered pairs.

The headroom for a cross $A_i \times A_j$ is the total number of pairs belonging to the cross that are \emph{incorrectly} ordered; we call this score \emph{number of incorrect pairs}. Consequently, if we sum the number of incorrect pairs over all crosses and divide by $p*n$, where $p$ is the number of positive examples and $n$ is the number of negative examples, we get 1- AUC, the headroom in the AUC. In this approach, a slice with a poor AUC will not be emphasized if it has a small size; we take the size of the slice into account, because we would like to decompose the headroom. This is an alternative to the size normalization in Section~\ref{sec:normalized-aggregation}.

\begin{figure}[h]
\caption{AUC headroom defined as the number of incorrectly ordered pairs, computed for the logistic regression model on \textit{Relationship} slices. Most of the model's mistakes occur on pairs where the negative item is \textit{Husband}. Note that while Figure~\ref{fig:cross-auc-lending} visualizes the fraction of correctly ordered pairs, this figure visualizes the number of incorrectly ordered pairs, not divided by the total number of pairs.\label{fig:cross-headroom-census}}
\centering
\includegraphics[width=0.45\textwidth]{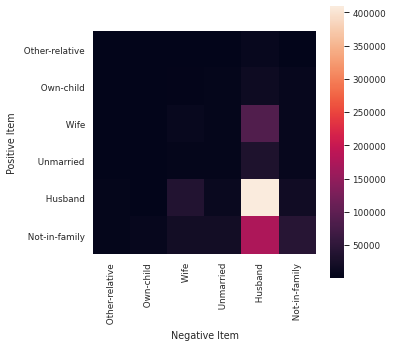}
\end{figure}

Figure~\ref{fig:cross-headroom-census} visualizes the headroom for slices in the Census dataset. The diagonal element corresponding to both positive and negative \emph{Husbands} in the pair contains only ~2\% of total pairs in the dataset. However, the model achieves an AUC score of 0.76 (compared to 0.88 overall). As a result most of the headroom for improving overall AUC on the dataset lies in that slice.

We generate a similar figure for Lending Club (omitted due to space constraints). We find that the the top right corner element in Figure~\ref{fig:cross-auc-lending} with near zero AUC only contains ~1\% of all pairs in the dataset. The third diagonal element corresponding to \emph{Grade C} contains ~9\% of all pairs in the dataset. Despite the model achieving AUC of 0.56 on the latter, there are still 4 times as many incorrectly ordered pairs in that slice than in the slice with near zero AUC. Improving model performance on the larger slice has a larger potential to increase overall AUC on the dataset.

% \begin{figure}[h]
% \caption{AUC headroom defined as the number of incorrectly ordered pairs, computed for the logistic regression model on \textit{Loan Grade} slices. Most of the model's mistakes occur on pairs involving \textit{Grade C} loans. Note that while Figure~\ref{fig:cross-auc-lending} visualizes the fraction of correctly ordered pairs, this figure visualizes the number of incorrectly ordered pairs, not divided by the total number of pairs.}
% \label{fig:cross-headroom-lending}
% \centering
% \includegraphics[width=0.5\textwidth]{Figures/lending-headroom-logistic.png}
% \end{figure}

\subsection{Automatically Segmenting AUC}
\label{sec:decision-boundary}

In this section, we automate the AUC segmentation analysis from Section~\ref{sec:normalized-aggregation}. Using a decision tree, we scale to a large number of slicing dimensions. 

\subsubsection{Segmentation Methodology}
Creating and labeling the Cartesian product of the evaluation set is costly in terms of runtime and memory. Hence, the positive and negative examples are sampled before pairing to reduce the cost. After constructing pairs out of roughly 10K positive and negative examples via Cartesian product, a label of $1$ is assigned to every correctly ordered pair (pairs where the positive example has a higher score  than the negative example) and $0$ to incorrectly-ordered pairs. A shallow decision tree is fit on this data similar to Example Attribution Segmentation in Section \ref{sec:segmentation}. In contrast to Section \ref{sec:segmentation}, each non-leaf node in this tree represents a split along a dimension from either the positive or negative examples.

\subsubsection{Results}
% In the next paragraph. Features lack positive and negative features.
Applying segmentation on pair attributions allows us to automatically uncover data slices where the model is having problem (or an easier time) separating negative and positive examples. Table \ref{table:pair_slices} and Figure \ref{fig:pair-segment-lending} show pairwise AUC attributions between  positive and negative segments for the Census and Lending Club models respectively. As can be seen, two different groups of segments may have drastically different AUC attributions. For example, the Lending Club model had a hard time correctly ordering defaulted, 36-month term loans by people with a high FICO credit score and paid off loans by people who have a low FICO score. Conversely, the Census model had a near-perfect performance on ordering married people with more than 50K income and unmarried people with less than $4,533$ capital gain and less than 50K income.

%Figure \ref{fig:pair-segment} shows a tree-based segmentation on the Census dataset across all columns. The leaf node representing the data slice $(MaritalStatus\_neg = Married-civ-spouse) \land (EducationNum\_pos<10.5) \land (CapitalGain\_pos<5056)$ implies low separability. In other words, the model is not very successful at correctly ordering negative examples in $(MaritalStatus = Married-civ-spouse)$ slice and the positive examples in the $(EducationNum<10.5) \land (CapitalGain<5056)$ slice. Conversely, negative examples in the $(MaritalStatus \neq Married-civ-spouse)$ slice can be separated easily from all other positive examples.
% If there were only examples from these slices, the model would have a low AUC score.

% \begin{figure*}[h]
% \caption{Tree-based segmentation of pairwise Census data. Negative features are suffixed by `\_neg' and positive features by `\_pos'.}
% \label{fig:pair-segment}
% \centering
% \includegraphics[width=\textwidth]{Figures/pairwise-segmentation-census.png}
% \end{figure*}

\section{Conclusion and Future Work}
\label{sec:conclusion}

In this paper, we propose simple, efficient techniques to attribute AUC-ROC to examples and pairs of examples. Analyzing these attributions demonstrates how the performance of a model varies across slices of data, often indicating that the model's performance would not be robust to changes in the mix/proportion of examples across these segments. We show that there are segments where simple models outperform complex ones. We also categorize pairs of examples that the model has trouble `separating'. We hope that these tools guide the ML developer in the process of improving the aggregate AUC.

For future work, one could extend this approach to other non-linear metrics such as the F-score, or the AUC  approach to multiclass and multilabel classifiers. One could also investigate how to distribute the implementations of the attribution and segmentation techniques; this seems plausible because the attribution technique relies on manifesting pairs of examples that can be implemented via distributed joins, and the segmenting technique could benefit from a distributed implementation of a decision tree classifier such as~\cite{xgboost}. 

% \todobesim{pairwise segmentation could be costly in terms of memory and runtime given the size of the cartesian produdct dataset.}

%\section{Experiments}
%\input{experiments}

\bibliographystyle{unsrtnat}
\bibliography{sample-base}

\clearpage

% \appendix{}
% \label{appendix}
% \input{appendix}

% \appendix

% \section{Appendix}

% \begin{figure}[h]
% \caption{}
% \centering
% \includegraphics[width=0.5\textwidth]{Figures/census-gini-corr-auc.png}
% \end{figure}
% \todo{depending on room, either move to main section or leave in appendix. add one for loss, change y-axis to normalized}

\end{document}